\documentclass[sigconf,screen]{acmart}

\graphicspath{{figs/}}
\usepackage{microtype} 
\usepackage{hyperref}
\usepackage{balance}
\usepackage{enumitem}

\AtBeginDocument{%
  \providecommand\BibTeX{{%
    \normalfont B\kern-0.5em{\scshape i\kern-0.25em b}\kern-0.8em\TeX}}}

\renewenvironment{quotation}{%
  \list{}{%
    \leftmargin0.5cm   
    \rightmargin\leftmargin
  }
  \item\relax
}
{\endlist}
\setlist{leftmargin=5mm}

\copyrightyear{2021}
\acmYear{2021}
\setcopyright{acmcopyright}\acmConference[UIST '21]{The 34th Annual ACM Symposium on User Interface Software and Technology}{October 10--14, 2021}{Virtual Event, USA}
\acmBooktitle{The 34th Annual ACM Symposium on User Interface Software and Technology (UIST '21), October 10--14, 2021, Virtual Event, USA}
\acmPrice{15.00}
\acmDOI{10.1145/3472749.3474773}
\acmISBN{978-1-4503-8635-7/21/10}

\begin{document}

\title{Situated Live Programming for Human-Robot Collaboration}


\author{Emmanuel Senft}
\email{esenft@wisc.edu}
\affiliation{%
  \institution{University of Wisconsin--Madison}
  \city{Madison}
  \state{Wisconsin}
  \country{USA}
}
\author{Michael Hagenow}
\affiliation{%
  \institution{University of Wisconsin--Madison}
  \city{Madison}
  \state{Wisconsin}
  \country{USA}
}
\author{Robert Radwin}
\affiliation{%
  \institution{University of Wisconsin--Madison}
  \city{Madison}
  \state{Wisconsin}
  \country{USA}
}
\author{Michael Zinn}
\affiliation{%
  \institution{University of Wisconsin--Madison}
  \city{Madison}
  \state{Wisconsin}
  \country{USA}
}
\author{Michael Gleicher}
\affiliation{%
  \institution{University of Wisconsin--Madison}
  \city{Madison}
  \state{Wisconsin}
  \country{USA}
}
\author{Bilge Mutlu}
\affiliation{%
  \institution{University of Wisconsin--Madison}
  \city{Madison}
  \state{Wisconsin}
  \country{USA}
}


\begin{abstract}
We present situated live programming for human-robot collaboration, an approach that enables users with limited programming experience to program collaborative applications for human-robot interaction.
Allowing end users, such as shop floor workers, to program collaborative robot themselves would make it easy to ``retask'' robots from one process to another, facilitating their adoption by small and medium enterprises.
Our approach builds on the paradigm of trigger-action programming (TAP) by allowing end users to create rich interactions through simple trigger-action pairings.
It enables end users to \emph{iteratively} create, edit, and refine a reactive robot program while executing partial programs.
This live programming approach enables the user to utilize the task space and objects by incrementally specifying situated trigger-action pairs, substantially lowering the barrier to entry for programming or reprogramming robots for collaboration. We instantiate situated live programming in an authoring system where users can create trigger-action programs by annotating an augmented video feed from the robot's perspective and assign robot actions to trigger conditions. We evaluated this system in a study where participants ($n=10$) developed robot programs for solving collaborative light-manufacturing tasks. Results showed that users with little programming experience were able to program HRC tasks in an interactive fashion and our situated live programming approach further supported individualized strategies and workflows. We conclude by discussing opportunities and limitations of the proposed approach, our system implementation, and our study and discuss a roadmap for expanding this approach to a broader range of tasks and applications.
\end{abstract}



\begin{teaserfigure}
  \centering
  \includegraphics[width=1\linewidth]{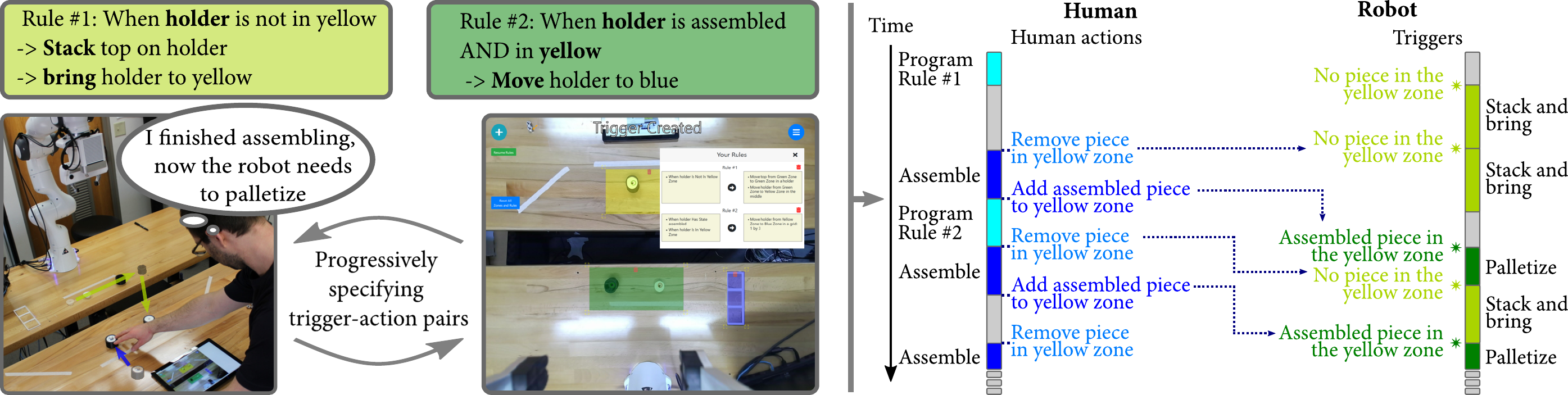}
  \Description[Teaser figure presenting the general idea behind Situated Live Programming for Human Robot Collaboration.]
    {
    Left: Left pane shows a photo of the workspace with the robot interacting with the user. There are multiple top and holder pieces that need to be assembled together. A thought bubble from the user says ``I finished assembling, now the robot needs to palletize''. The right pane shows a picture of the interface with the robot's view, zones created by the user, and two trigger-action pairs: (1) ``When holder is not in yellow, stack top on holder, bring holder to yellow'' (2) ``When holder is assembled and in yellow, move holder to blue''.
    Right: Interaction flow as a timeline with human actions on the left and robot actions on the right. When the human modifies the workspace, the rules trigger an action by the robot. The final human timeline alternates between assembling pieces and programming new trigger-action pairs.
    }
  \caption{This paper presents an end-user programming approach for collaborative robots, called \textit{situated live programming (SLP)}, to perform tasks that require responsiveness and coordination. \textit{Left:} The user incrementally specifies an assembly task as trigger-action pairs. 
  \textit{Right:} An example collaborative program created using SLP. The triggers, human actions (blue), and robot actions (green) are mapped on a task timeline. 
  }
  \label{fig:teaser}
\end{teaserfigure}

\maketitle

\section{Introduction}
Collaborative robots (cobots) are flexible robot platforms designed to perform physical tasks in the presence of people.
They are finding increasingly widespread adoption across industries, from manufacturing to healthcare \cite{bauer2008human,chandrasekaran2015human}. A key facilitator of this adoption is the promise of flexibility: the premise that cobots are easy to ``retask'' from one process to another. However, in practice, this adoption is restricted to isolated, solitary tasks \cite{michaelis2020collaborative}.
While current platforms make some tasks easy to program, considerable expertise is required for tasks where the robot must coordinate with human collaborators or other systems in the environment, such as a conveyor belt.
Solutions that achieve such coordination and integration require expert implementation or precise modeling of the process \cite[e.g.,][]{gombolay2015decision,tsarouchi2016human} and thus are not suitable for use by non-experts such as shop floor workers.
Cobots usually provide end-user programming methods, such as programming by demonstration (PbD), but these methods do not support specifying coordination. 
New end-user programming methods and tools are needed 
that enable end users to specify complex tasks with coordination.

In this paper, we present an end-user programming approach called \emph{situated live programming for human-robot collaboration}  (abbreviated ``SLP'' or ``SLP for HRC'' in the remainder of the paper) that enables users with little programming experience to program robots to perform tasks that require responsiveness and coordination. 
Our work addresses the challenge of integrating collaborative robots into an existing task environment with human workers and other systems by making it easier to program responsive robot behaviors. 
SLP builds on trigger-action programming (TAP), a paradigm where users build programs by specifying trigger-action pairs. 
In SLP, robot operators specify triggers and actions in the environment, extending TAP to be used incrementally for robot programming.
For example, operators can outline a work area, such as a workbench or a container, and specify an action that the robot can take when objects of interest are placed on this area by drawing on a tablet (see Figure \ref{fig:teaser}). The appearance of expected objects triggers the robot to take the specified action. 
Defining such trigger-action pairs in a sequence enables the programming of complex tasks. The ability to define the consequences of the actions of human collaborators and other systems as triggers facilitates the integration of the robot's actions with the actions of other agents and systems.
The incremental specification of simple triggers and actions, along with the feedback provided by the execution of these specifications, enables the user to make progress toward the goal program without knowing what the complete solution must involve.
By enabling end users to easily specify responsive robot programs that integrate with the environment, including human co-workers, SLP can help overcome a key bottleneck in the adoption of collaborative robots. 

We implemented a prototype SLP system contextualized in a light manufacturing operation involving a cobot and a human worker. Our system integrates an interface where users annotate an augmented video feed from a camera mounted on the robot to specify triggers in the physical world. We performed an exploratory study where 10 participants interacted with our system to solve tasks such as sorting, kitting (packing various parts into a ``kit''), and collaborative assembly with the robot. The study showed that users with little programming experience were able to program HRC tasks in an interactive fashion and that SLP further supported individualized strategies. Furthermore, participants reported how live programming enabled them to discover solutions to the task and create a mental model of the interaction. 
The results of the study show the promise of the SLP approach: non-experts were able to create complex programs that responded the robot's environment, including human collaborators.

Our work addresses the challenge of enabling non-experts to program robots for collaborative tasks. Specifically, the contributions of this work include:
\begin{enumerate}
    \item The formulation of \textit{situated live programming (SLP)} as an extension of current TAP methods to support \textit{incremental} programming of rules in \textit{collaborative tasks};
    \item The application of TAP to collaborative robotics;
    \item An open-source SLP-based authoring tool;\footnote{\url{https://osf.io/s48te/?view_only=6daa6020059c4db4838cca193ab58df4}}
    \item A user study exploring how users with little programming experience can reason about and create trigger-action programs for collaborative tasks;
    \item Discussion of the limitations of SLP and a roadmap to advance TAP-based programming in human-robot collaboration.
\end{enumerate}

\section{Related Work}
    \subsection{Human-Robot Collaboration}

Human-Robot Collaboration (HRC) refers to situations where a team of human(s) and robot(s) solves tasks together. HRC has been applied to a wide range of applications from manufacturing to social robotics \cite{bauer2008human,chandrasekaran2015human}. Programming methods for robot collaboration need to address two challenges: grounding actions in the real world and coordinating agents. 
Methods that enable end users to address the first challenge include visual programming or programming by demonstration \cite{rossano2013easy}. 
In these methods, programming is completed prior to use in the world and may require users to demonstrate the task or handle coding functions such as callbacks, loops, and conditional branching. Teach pendants (such as the UR Polyscope\footnote{\url{https://www.universal-robots.com/blog/tags/polyscope-56/}}), the standard input for all industrial collaborative robots, offer a satisfactory solution when there is little to no variation in the environment. However, such low-level methods relying on programming by demonstration are time consuming and error prone \cite{gao2019pati} and provide limited robustness to changes in initial conditions (e.g., the starting position of objects).
 More recent work has explored ways to make this programming easier, for example, by focusing only on sequencing of high-level actions \cite{steinmetz2018razer,steinmetz2019intuitive}. Other work has addressed this grounding challenge by allowing users to ground actions in a virtual representation of the world \cite{huang2017code3,huang2020vipo}, using monitor-based augmented reality (AR) \cite{kapinus2019spatially}, head-mounted AR displays \cite{bambusek2019combining}, or by direct annotation of a retroprojected workspace \cite{gao2019pati}. However, most of these approaches are focused on specifying behaviors to perform navigation or manipulation rather than specifying robot behaviors for collaboration. These approaches focus on \textit{what} to do, and \textit{how} to do it, but scarcely on \textit{when} to do it. Furthermore, these methods aim to create a fixed autonomous behavior, often using a \textit{program-compile-evaluate} workflow, which can be inconvenient as it makes testing and refining behaviors time consuming.

Other programming methods have approached HRC from a planning or scheduling perspective. In these methods, all agents (humans and robots) and the task are precisely modeled and a scheduling algorithm optimizes the action allocation to each agent to reduce cost and/or time and handle human variations \cite{gombolay2018fast}. The models in these approaches often use abstract representations of objects and require engineers as designers. Most recently, Schoen et al. \cite{schoen2020authr} explored how end users could create tasks themselves, define the capabilities of agents, and let the algorithm optimize the allocation of actions in known environments. However, this method relies on exact timing of agents and open-loop interaction, which means that if an agent is slower than expected or executes an unanticipated action, the interaction would break down.

In summary, intuitive methods for robot programming (e.g., programming by demonstration) \cite{rossano2013easy} are not designed for interaction, and scheduling methods require either expert users or accurate models of each agent. These methods are either not designed for creating interactive behaviors, or not designed for novice users and consequently fail to provide non-programmers ways to design programs for collaborative robots.

\subsection{Trigger-Action Programming}

Trigger-Action Programming (TAP) is a programming paradigm based on trigger-action pairs. TAP is mostly used in home automation and Internet of Things (IoT), with one of the most popular application being the IFTTT (If This Then That).\footnote{\url{https://ifttt.com/}} IFTTT allow users to create rules associating one action to a specific trigger: for example, ``If `it is past 6pm' then `turn the light on','' As shown by Ur et al. \cite{ur2016trigger}, IFTTT is widely used by end users to create rules for home automation or app synchronization (for example sending a message on an app when an email is received). Similar products exist in workflow automation such as Microsoft Power Automate\footnote{\url{https://flow.microsoft.com/en-us/}} or the TASKER android app.\footnote{\url{https://tasker.joaoapps.com/}}

Triggers can be either events (``it starts raining'') or states (``it is raining'').
This distinction is important as the type of trigger creates different expectations for users \cite{huang2015supporting}, even though there can be a conversion between event and states. While IFTTT uses trigger-action pairs composed of a single trigger and a single action, other work has explored more complex pairs with more than one condition in the trigger and more than one action \cite{ur2014practical}.

In the last five to ten years, research has explored multiple aspects of TAP, for example the importance of context when creating trigger-action programs \cite{ghiani2017personalization}, how people create mental models of trigger-action programs \cite{huang2015supporting} , and how people discover and address bugs in programs \cite{brackenbury2019users}. Other work has explored how to synthesize trigger-action programs from behaviors traces \cite{zhang2020trace2tap} or repair them using formal logic \cite{zhang2019autotap} which changes the way trigger-action programs can be created or represented.

TAP has also been used recently in social robotics \cite{leonardi2019trigger} and to integrate robots in IoT environments \cite{huang2020vipo}, but to our knowledge, TAP has never been applied to HRC. As shown with IFTTT, TAP applications have focused on relatively simple rules with known sensors and actuators. However, to be applicable to HRC, where objects can appear, move, or be transformed at runtime by other agents, TAP needs to be adapted and extended.

\subsection{Robot Live Programming}
\label{sec:live}
Live programming refers to the ability to update a program while it is running \cite{tanimoto2013perspective}. Live coding has seen a widespread use in music generation as a way to compose and generate music live \cite{collins2003live}. In classic programming, it often refers to a program displaying the impacts of parameters while the user manipulates them \emph{live}. The main goal of live programming is to move away from the traditional edit-compile-run loop and towards a much tighter loop, allowing developers to get rapid feedback on their programs, test their assumptions early in the development process, simplify the debugging process, and support learning. For example, McNerney \cite{mcnerney2004turtles} presents numerous tangible programming methods possessing both a situated aspect and a live aspect to support teaching computing and scientific concepts to children. Notable examples include a train track that could be annotated to change the train's \textit{program} (e.g., to define or move stops) and a drawing application where the drawing path could be changed partway through the task. 

Live programming has seen applications both in programming languages and integrated development environments \cite{kubelka2018road}. In 1990, Tanimoto \cite{tanimoto1990viva} proposed four levels of liveliness (informative, significant, responsive, and live), which range from having to restart a program to applying changes to a live application reacting in real-time. In 2013, Tanimoto \cite{tanimoto2013perspective} extended these four initial levels to cover programs that can predict the user's intents. Recently, the intersection of live programming and language synthesis has seen a high interest in the community, for example how part of a program could be generated live using snippets of data \cite{ferdowsifard2020small} or how synthetic data can be created at design time to evaluate a loop \cite{lerner2020focused}. 

In the context of robotics, live programming refers to the ability to update the code of an active robot while it is interacting with the real world. However, previous work has not applied live programming to human robot collaboration. Existing examples focus on robots for education \cite{shin2014visual} and manufacturing robots using text-based programming within nested state machines \cite{campusano2017live}.

Our work leverages features from HRC, TAP, and live programming. We provide further detail on these features in Section 3 and compare them to existing methods in Section 3.3.

\section{Situated Live Programming for Human-Robot Interaction}
This paper introduces Situated Live Programming for Human-Robot Collaboration. SLP builds on  trigger-action programming (TAP) by creating interactive robot behaviors emerging from the interconnection of multiple trigger-action pairs. SLP extends TAP in two ways to make it applicable to HRC. 

\paragraph{Situating TAP in the physical world.} Typically TAP includes manipulating abstract symbols to refer to states, events, data (e.g., emails and messages), or actions related to known IoT devices, for example by using buttons on a smartphone application to synchronize a defined motion detector and a defined light bulb. However, HRC environments are often dynamic and objects' location, count, and states can change over time. To address this characteristic of HRC, we situate and ground TAP in the physical world by using a live camera feed as the authoring background.

\paragraph{Live Programming.} TAP is generally composed of independent trigger-action pairs. With SLP, we create an interactive program by combining multiple pairs. This combination of distinct elements of the program allows for live programming. Instead of building a behavior first and then deploying it on a robot to test it, SLP allows for incremental solutions toward a goal. It also allows for complex tasks to be solved as a sequence of simpler steps.

\subsection{Situated TAP for HRC}

HRC requires two or more agents to coordinate their actions to complete the task at hand. For example, in an assembly task, a human might place bolts in holes and let the robot tighten them.
Successful collaboration for such interactions requires the robot program to appropriately define when, where, and how the robot should act. TAP is a natural formalism to represent such robot reactions to human actions and would allow workers to define interactive robot behaviors themselves. The human can define what conditions should trigger a specific action. For example, for the bolt tightening task, the worker can define the area where the bolts will be placed and create the following trigger-action pair: if ``bolt in the area is not tightened,'' then ``tighten it.'' Assuming that a robot can detect loose bolts, this type of instruction would ensure that only loose bolts would be tightened and would also be robust to the order in which the worker places the bolts.


\subsection{Situated Live Programming}

As a trigger-action program consists of a number of trigger-action pairs, new pairs can be easily added while the robot is running, and thus TAP also supports live programming.
Live programming presents a number of desirable qualities for programming HRC. In contrast to traditional programming where a user specifies the full behavior (sequence, branches, and loop) prior to execution, with SLP the user does not need a full mental model of the task to start programming the robot. Instead, the user can start by programming the first step of the interaction, and then progressively build the full program. At each step, the user can define new trigger-action pairs based on the current state of the environment. 

Live programming also allows users to handle errors and edge cases only if they arise. While a robot with a static program needs to be robust to any possible error or edge case, a robot in a live programming paradigm can have its program corrected at runtime, without needing to address all errors ahead of time. For example, if a worker realizes that a robot action could fail if the robot is too fast for them, then they can update the condition of the trigger-action pair to make sure that this edge case is covered.

SLP inherit the benefits of both live programming (i.e., allowing programmers to get feedback early and often to make the programming process easier to learn and to use \cite{kubelka2018road}) and situated programming (i.e., easily grounding the program in the current state of the environment).

\subsection{Comparison to Existing Methods}
Although situated and live programming has been explored in prior work (see Section \ref{sec:live}), particularly in education, these methods would require significant changes to be applicable to HRC scenarios, for example, allowing users to refer to objects in the space, parameterize actions, and react to human inputs and state changes provoked by outside agents. Our vision of SLP for HRC is to support novice users in developing \textit{interactive} robot programs for human-robot collaboration, an aspect not explored in past systems.

SLP differs from traditional TAP in three key ways. First, the robot program is defined by how multiple TAP rules interact with each other and react to human actions in the workspace. Second, SLP allows users to layer TAP rules on top of each other to iteratively create a robot behavior. Finally, SLP handles more complex states, where robot actions need to be parameterized and contextualized, potentially taking as parameter multiple objects not visible at design time or needing information to disambiguate between multiple instances of the same object type. Traditionally, TAP works with known sensors and known actuators and matches fully specified actions to conditions on sensors. When working with data (e.g., forwarding emails), TAP handles new instances of objects, but the context is most of the time not ambiguous. 

Finally, compared to end-user programming methods such as task-level programming \cite{steinmetz2018razer} or programming by demonstration, SLP aims to both provide flexible programs (i.e., programs handling different starting and ending states) and define when actions should be executed in a reactive fashion. Furthermore, SLP leverages TAP as primitives to incrementally program robotic workflows instead of approaching programming as a batch process.

\section{Implemented Prototype}
\label{sec:system}
We implemented an initial prototype of SLP contextualized in light-manufacturing operations. This system is centered around a collaborative robot with a camera attached to the end-effector and a tablet where users can program the robot by annotating the video feed and creating trigger-action pairs.

\subsection{Apparatus}

\begin{figure}[!t]
  \centering
  \includegraphics[width=1\linewidth]{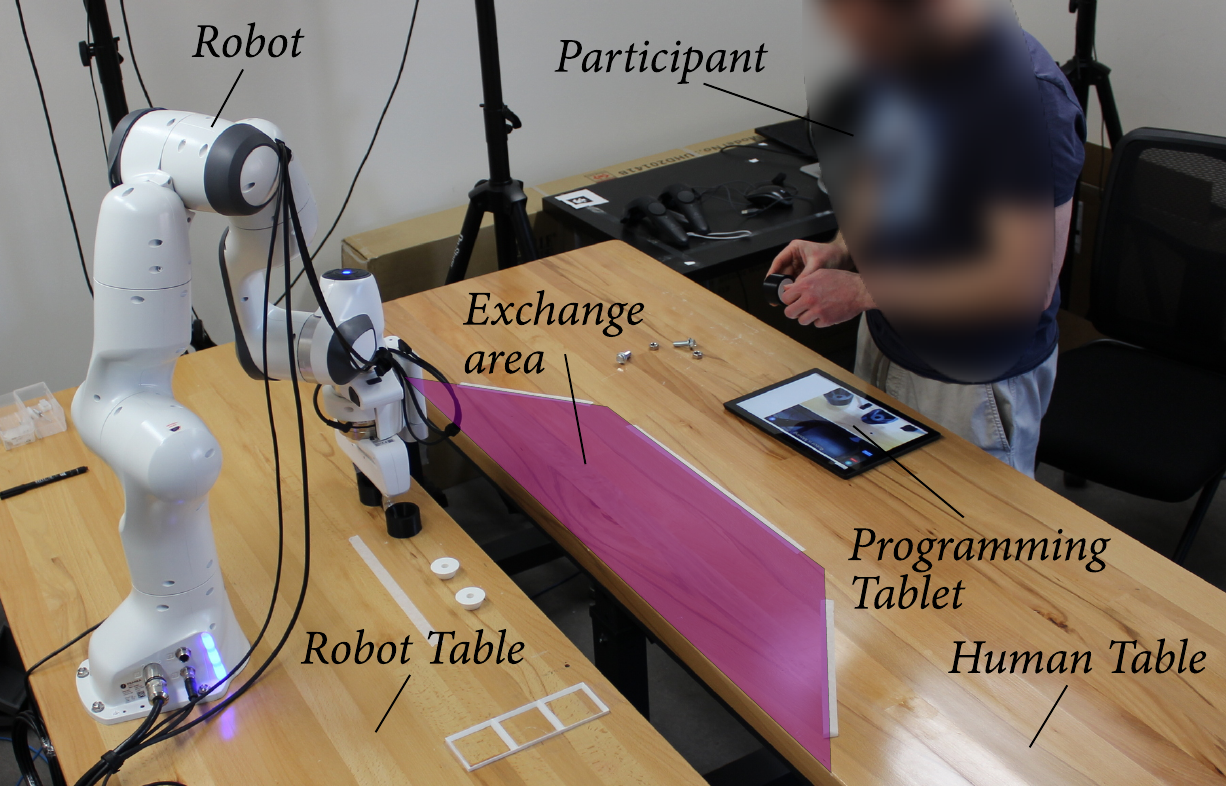}
  \Description[Annotated picture of the workspace showing a collaboration between the robot and the user.]
    {
    The workspace is composed of two tables, the robot, and the user. Parts of the assembly are sitting on the robot table, next to an empty palletizing area. The human table has a labeled work area, which is outside the range of the robot and delineated with tape, as well as a tablet for programming.
    }
  \caption{Workspace used for the system. Both the robot and the human had their own table with a exchange area delineated by masking tape marking the robot reach.}
  \label{fig:workspace}
\end{figure}

Our implementation integrates a collaborative robot, Franka Emika Panda\footnote{\url{https://www.franka.de/technology/}}, outfitted with an ATI Axia80-M20 6-axis force torque sensor and a Microsoft Azure Kinect providing an RGB-D image. The camera is placed on the robot end-effector to allow for the greatest flexibility in camera navigation while maintaining an inherent registration between the vision system and the robot. 
The software components of the system communicate using ROS \cite{quigley2009ros} with logic nodes implemented in Python. The graphical user interface is developed in Javascript with React and runs on a Microsoft Surface Pro tablet. The low-level robot control operates at 1000Hz and is implemented in C++.\footnote{Open-source code for our system implementation is available at \url{https://osf.io/s48te/?view_only=6daa6020059c4db4838cca193ab58df4}.} 
The robot is placed on a table, with the human sitting or standing on the other side of a second table, outside of the robot's reach (see Figure \ref{fig:workspace}).

\subsection{User Interface}

\begin{figure*}[t]
  \centering
  \includegraphics[width=1\linewidth]{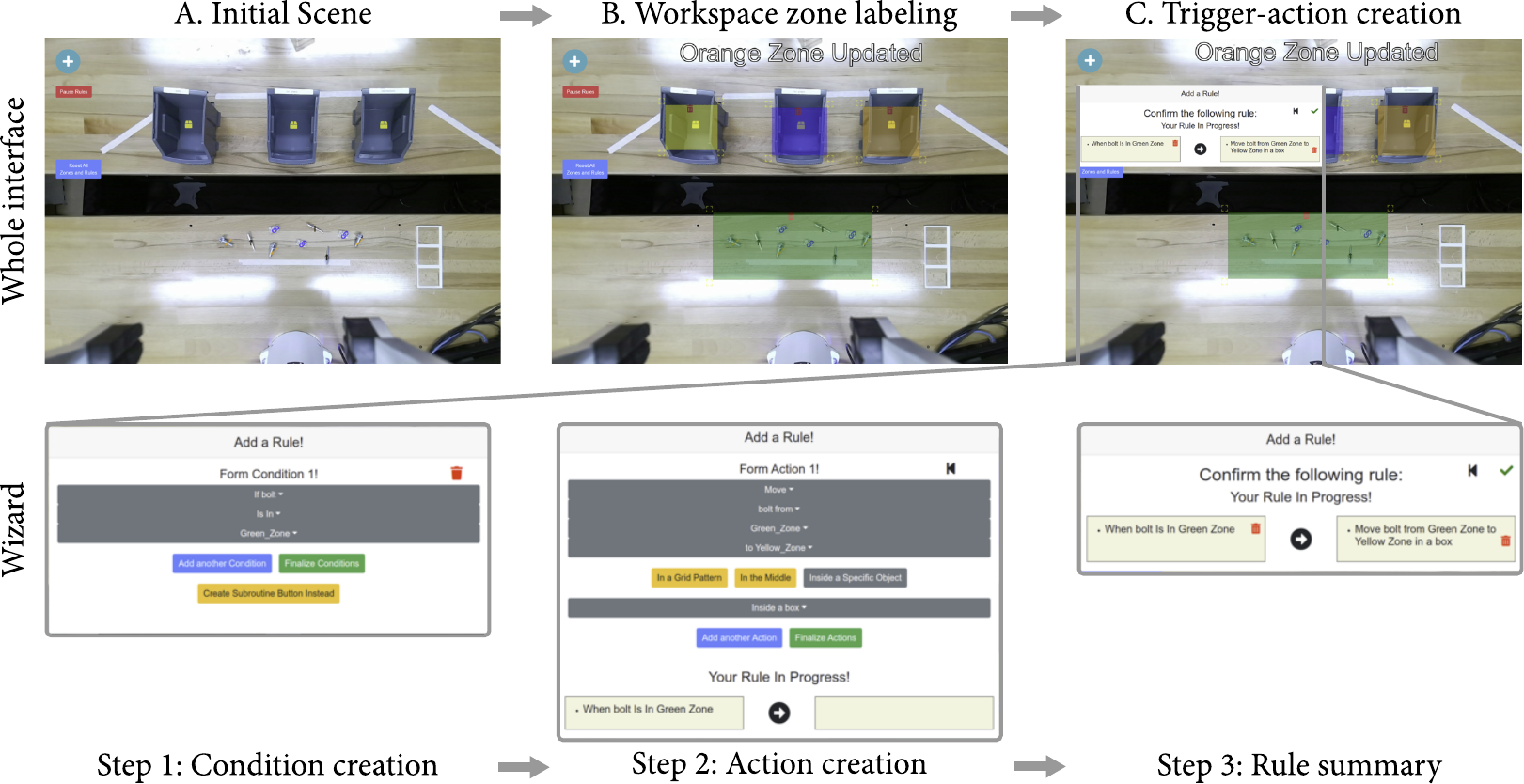}
    \Description[Step-by-step example of using the interface to create a trigger-action pair.]
    {
    Top row is labeled ``whole interface''. The first image is labeled ``A. Initial scene'' and shows bolts, connectors, and fasteners on the robot table and three bins on the human table. The second image is labeled ``B Workspace zone labeling'' and shows four zones of different colors, one around all the objects and one for each box. The third image is labeled ``C. Trigger-action creation'' and shows the same image with the colored zones, but with a menu summarizing the current trigger-action creation (When bolt is in green zone, move bolt from green zone to yellow zone in a box). A callout from this menu extends to the three images in the bottom row.
    The bottom row is labeled ``Wizard'' and consists of close-up shots of the interface during rule creation. The first image is labeled ``Step 1: condition creation'' and shows a menu with three dropdowns to specify a condition (if bolt - is in - green zone), with buttons to validate or add more conditions. The second image is labeled ``Step 2: action creation'' and shows a menu with five dropdowns to specify an action (move, bolt from, green zone, to yellow zone, inside a box), with buttons to change the placement pattern before the last dropdown. The third image is labeled ``Step 3: Rule summary'' and displays a summary of the trigger-action pair with a list of conditions on the left (when bolt is in green zone) and a list of actions on the right (move bolt from green zone to yellow zone in a box). 
    Images within rows are connected with arrows from left to right.
    }
  \caption{Interface workflow while creating a trigger-action pair to move all of the bolts from the green zone to the storage box in the yellow zone. The top row represents the higher level flow: A. objects are on the robot's table and the boxes on the human's table, with icons showing that the robot detected them. B. The user starts by labeling the interesting areas: one large zone for all the objects and one smaller zone per box. C. The user creates trigger-action pairs. The bottom row presents a focus on the wizard tool to create a trigger-action pair: Step 1: creating the condition: ``if a bolt is in the green zone.'' Step 2: Selecting the action: ``move bolt from the green to the yellow zone, in a box.'' Step 3: Verifying and confirming the pair before adding it to the robot program.}
  \label{fig:ui_flow}
\end{figure*}

The user interface is centered around a robot-centric (camera in hand) video-feed which is overlayed with icons showing detected objects (see Figure \ref{fig:ui_flow}). Users can annotate this augmented feed to label zones of interest in the workspace. Zones are identified by a color from an accessible color palette and can be resized, moved, or deleted at anytime. Users can create trigger-action pairs (associating conditions and actions) through a wizard interface. 

With SLP, triggers and actions are exposed to the user as high-level constructs to simplify the programming. For example, an action would move an object from a starting zone to a destination zone. Additionally, both triggers and actions need to be generalizable to multiple instances of an object (e.g., to move any object, current or future, of a specific type from a starting zone to a destination zone). Consequently both triggers and actions should work at the object category level (e.g., \textit{any bolt}) instead of the instance level (e.g., \textit{bolt \#1}). An example of a trigger-action pair shown in Figure \ref{fig:ui_flow} is ``When [a] bolt is in [the] Green zone, Move [that] bolt from [the] Green Zone to [the] Yellow Zone, in a box.'' This specification will have the robot move all the bolts from the green zone to the box in the yellow zone.
The disambiguation of which specific instance of object needs to be manipulated is done at runtime by using objects satisfying both the trigger conditions and the action parameters.

Users can create trigger-action pairs at any point in the interaction. To create a trigger-action pair, users first specify a number of conditions using dropdowns referring to objects detected in the workspace (see Figure \ref{fig:ui_flow}, Step 1). Conditions are structured with an object category (e.g., \textit{bolt}), a condition (e.g., \textit{is in}), and condition parameter (e.g., \textit{Green zone)}. Our implementation supported the following triggers:
\begin{itemize}
    \item The presence of an object in a zone (e.g., bolt \textit{is in} Green zone);
    \item The absence of objects from a zone (e.g., bolt \textit{is not in} Red zone);
    \item The state of an object (e.g., holder \textit{has state} empty).
\end{itemize}
All the conditions are linked with an ``AND'' operator, which means that only if all the conditions in the trigger are true, the actions will be executed. We decided not to include the ``NOT'' and ``OR'' operators to keep the end-user programming method as simple as possible. In addition to specifying conditions, users could also dynamically create buttons to manually trigger action sets. 

After specifying the trigger, users can create associated actions. The only action exposed through the interface is a \textit{move} action which has four parameters:
\begin{itemize}
    \item Object category (e.g., move \textit{bolt});
    \item Starting zone (e.g., from \textit{Green Zone});
    \item Destination zone (e.g., to \textit{Yellow Zone});
    \item Placement pattern with three options:
    \begin{enumerate}
    \addtolength{\itemindent}{3mm}
        \item In a grid pattern (e.g., in a grid with \textit{1} column and \textit{3} rows);
        \item In the middle;
        \item Inside a specific object (e.g., in a \textit{box}).
    \end{enumerate}
\end{itemize}
As shown in Figure \ref{fig:ui_flow}, an example of fully parameterized move action is ``Move \textit{bolt} from the \textit{Green zone} to the \textit{Yellow zone}, in a \textit{box}''.
All these parameters are set through dropdowns that are dynamically populated with the objects in the workspace. This method for parameterizing the move action allows the robot to move objects between the tables, stack objects on top of each other, place objects in containers (e.g., box), and perform \emph{palletizing} actions. Of note, other actions are supported by the backend (e.g., push, tighten, and loosen) but were not exposed to the interface for simplicity.

In our trigger-action formulation, conditions are treated both as a trigger--if the condition is true, the action(s) will be executed--and as a filter to contextualize the objects used in the action. For example, in a pair such as ``if a holder has state empty'' then ``move holder from zone A to zone B, in the middle,'' only holders with the state \emph{empty} would be considered for the move action. This use of context is similar to verbal languages where a context is constructed in the sentence. 

\begin{figure}[!b]
  \centering
  \includegraphics[width=1\linewidth]{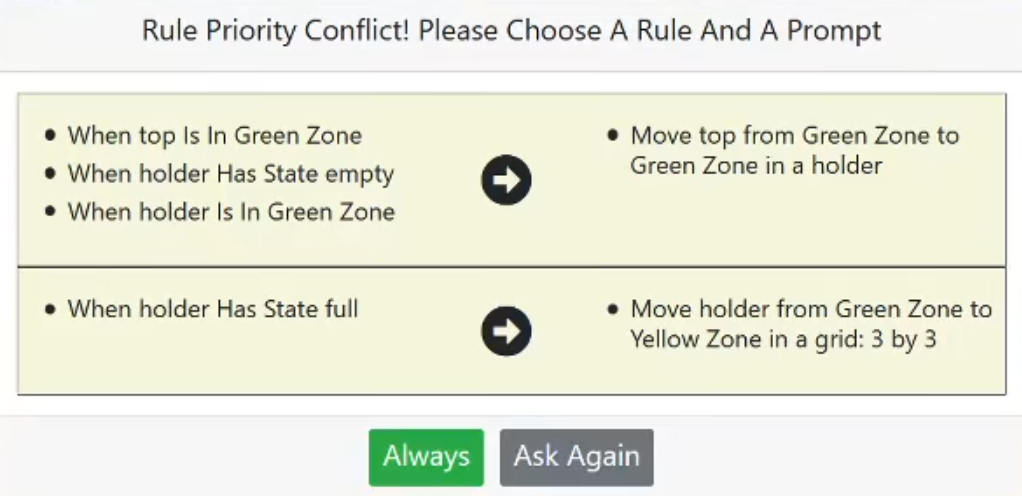}
    \Description[Priority conflict arbitration panel.]
    {
    The title of the panel is “Rule Priority Conflict! Please Choose A Rule And A Prompt”
    Then, two rules are displayed as two rows, with each rule composed of the trigger part on the left describing the conditions, the action part on the right, and an arrow connecting the triggers and actions. 
    At the bottom of the images, two buttons are labeled “Always” and “Ask Again.”
    }
  \caption{Interface pop-up showing a priority conflict. At runtime, the user can select which rule they want to execute and inform the robot whether or not they want this decision to be remembered.}
  \label{fig:priority}
\end{figure}

Users can also pause the robot execution if they desire to create multiple rules at the same time. Users also have access to a summary of their program where they can delete incorrect or obsolete trigger-action pairs. Finally, the interface can be used to answer priority conflicts at runtime. As shown in Figure \ref{fig:priority}, when more than one rule could be triggered, a pop-up prompts the user for which rule should be prioritized and whether future conflicts should be handled in a similar way. Priority conflict is a common occurence with TAP \cite{huang2015supporting,brackenbury2019users}, and using runtime query circumvents specifying all possible conflicts at design time.

Finally, a number of features have been included in the rule creation wizard to simplify the process:
\begin{enumerate}
    \item The dropdowns make use of context (such as previous conditions or actions) to autofill the default fields;
    \item Dropdowns are only filled with the objects currently in the workspace (allowing a large catalog of objects without crowding the dropdowns);
    \item Users can navigate in the wizard menu to change the number and parameters of both conditions and actions;
    \item Before creation of a pair, a summary panels allows to confirm the action, and delete undesired components.
\end{enumerate}  

\subsection{System}

\begin{figure}[!t]
  \centering
  \includegraphics[width=1\linewidth]{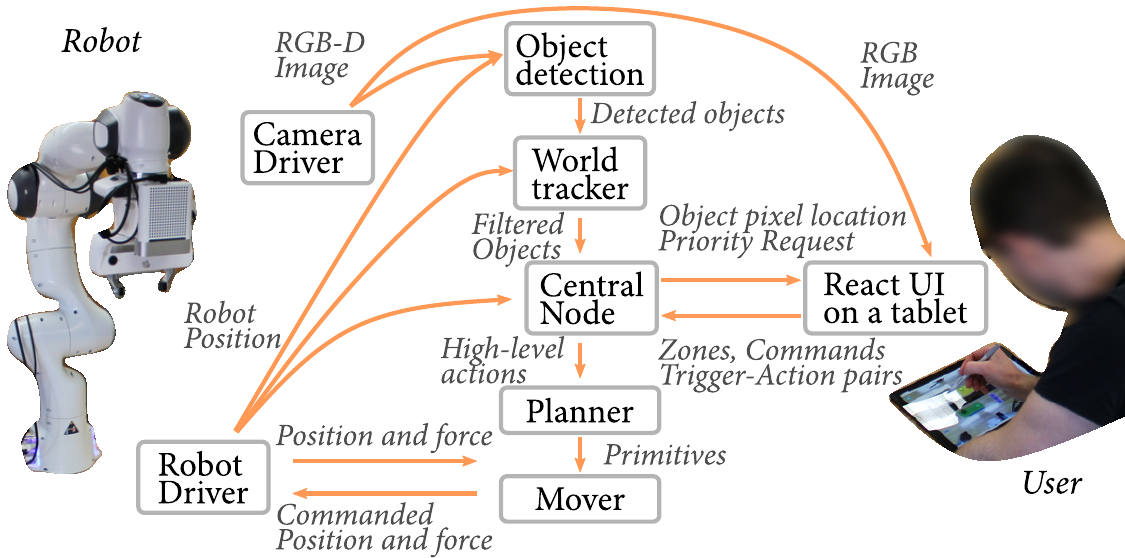}
    \Description[Flowchart of the system architecture.]
    {
    Picture of the robot on the left, one of the human on the right, and eight nodes in the middle connected by flow links. The flow goes from the ``Camera driver'' (robot side) and the ``React UI on a tablet'' (human side) nodes to the ``Robot Driver'' (robot side), and most of the nodes are connected to a central node.
    The flowchart has the following elements and connections:
    Camera Driver
    a. Sends RGB image to the node ``React UI on a tablet''
    b. Sends RGB-D image to the node ``Object detection''
    React UI on a tablet:
    a. Sends zones, trigger-action pairs, and commands to the ``Central node''
    Object detection
    a. Sends detected objects to the ``World tracker''
    World tracker
    a. Sends filtered object to the ``Central node''
    Central node
    a. Sends Object location on image and priority request to the node ``React UI on a tablet''
    b. Sends High-level actions to the ``Planner''
    Planner
    a. Sends Primitives to the ``Mover''
    Mover
    a. sends commanded position and force to ``Robot driver''
    Robot driver
    a. Sends positions and forces to the ``Mover''
    b. Sends robot position to the node ``Object detection'', the ``World tracker'', and the ``Central node''
    }
  \caption{High-level schematic of the system architecture.}
  \label{fig:arch}
\end{figure}

A simplified representation of the architecture controlling the robot is presented in Figure \ref{fig:arch}.

To act on the trigger-action pairs created by the users, the system needs an adequate representation of objects present in the workspace. Object recognition is achieved using Detectron2 \cite{wu2019detectron2}. A total of eight objects were manually labeled and used to retrain the network. Once detected in the image, objects are localized in 3D by using the depth map from the camera (or using the intersection with the table when in dead zones of the depth camera). To allow adequate grasping, the grasping orientation is aligned using principal component analysis with the principal axes of the object. A world state tracker filters the objects detected by the network to maintain a stable live representation of the state of the world.

A central node in the backend (named \emph{parser} in the sources) centralizes the information and communicates with the user interface. This parser keeps track of the zones and trigger-action pairs created by the user. At 2Hz, this parser publishes the position of detected objects and zones on the streamed image (according to the current camera location) and evaluates the triggers. For each trigger-action pair, if all the conditions are true and there is a candidate for the action series, then the pair is flagged as executable. As this trigger check is discrete (every 500 ms), multiple pairs could be flagged as executable in a single evaluation step. If this situation occurs, a priority request is sent to the user. In response to this request, the user can select which trigger-action pair they wish to see executed and specify whether this action should be selected again in the future (see the arbitration pop-up in Figure \ref{fig:priority}). As the robot can only execute one action at a time, the trigger evaluation is disabled when the robot is already moving.

When one pair should be executed, the actions are parameterized with the current state of the world, for example, by selecting which object instance should be moved. If multiple instances of an object satisfy the conditions, the system will execute the action on the first object detected by the object detection node. Then, the parameterized actions are forwarded to a planner that translates them into parameterized primitives (e.g., reaching a Cartesian position in space or opening the gripper) and sends them to an executor to make the robot move and manipulate objects in the workspace.

\section{Exploratory Study}

As a first evaluation of SLP, we designed an exploratory study to observe how participants would use SLP to construct interactive programs for HRC. The two goals of the study were to assess whether participants with little programming experience could reason about interactive programs as sets of trigger-action pairs and to explore what strategies participants would use with SLP.

\subsection{Participants}

We recruited 10 participants (6 males and 4 females), aged 20.9 years on average ($SD=1.2$), for an in-person study in the laboratory. The study protocol was approved by the University of Wisconsin--Madison Institutional Review Board. The participants were undergraduate (at least sophomore standing) and graduate students studying either Mechanical Engineering or Industrial and Systems Engineering . All participants had basic experience in programming (i.e., took at least one programming class at the university and/or had high school programming experience), but most of them were not expert programmers. Two participants reported that they rarely programmed, six sometimes, one often, and one very often. Similarly, participants had a low familiarity with robotics ($M=2.4$, $SD=0.8$ on a 1-5 scale) and home automation systems ($M=1.8$, $SD=.08$). The study lasted approximately 80 minutes.

\subsection{Study Method}

The experimenter introduced the robot and described the study to the participants. Then, participants read and signed a consent form, followed by a demographics questionnaire. Once the questionnaire was completed, participants were informed about required safety rules before starting the tasks. Participants completed four tasks (sorting, kitting, assembly, and assembly without live programming) in a fixed order that are detailed in the next section. Before each of the first three tasks, participants watched a video introducing them to the concepts related to the task\footnote{Instruction videos can be found at \url{https://osf.io/s48te/?view_only=6daa6020059c4db4838cca193ab58df4}} and were asked to verbally describe how they planned to solve the task. After completing the final task, participants responded to a questionnaire that included the System Usability Scale (SUS) \cite{brooke1996sus} and a scale of fluency in collaboration \cite{hoffman2019evaluating}. Finally, the experimenter conducted a semi-structured interview covering topics such as the system, the interface, structuring programs using TAP, or how the live programming approach differed from the non-live one used in the last task.

\subsection{Tasks}

\begin{figure}[!t]
  \centering
  \includegraphics[width=1\linewidth]{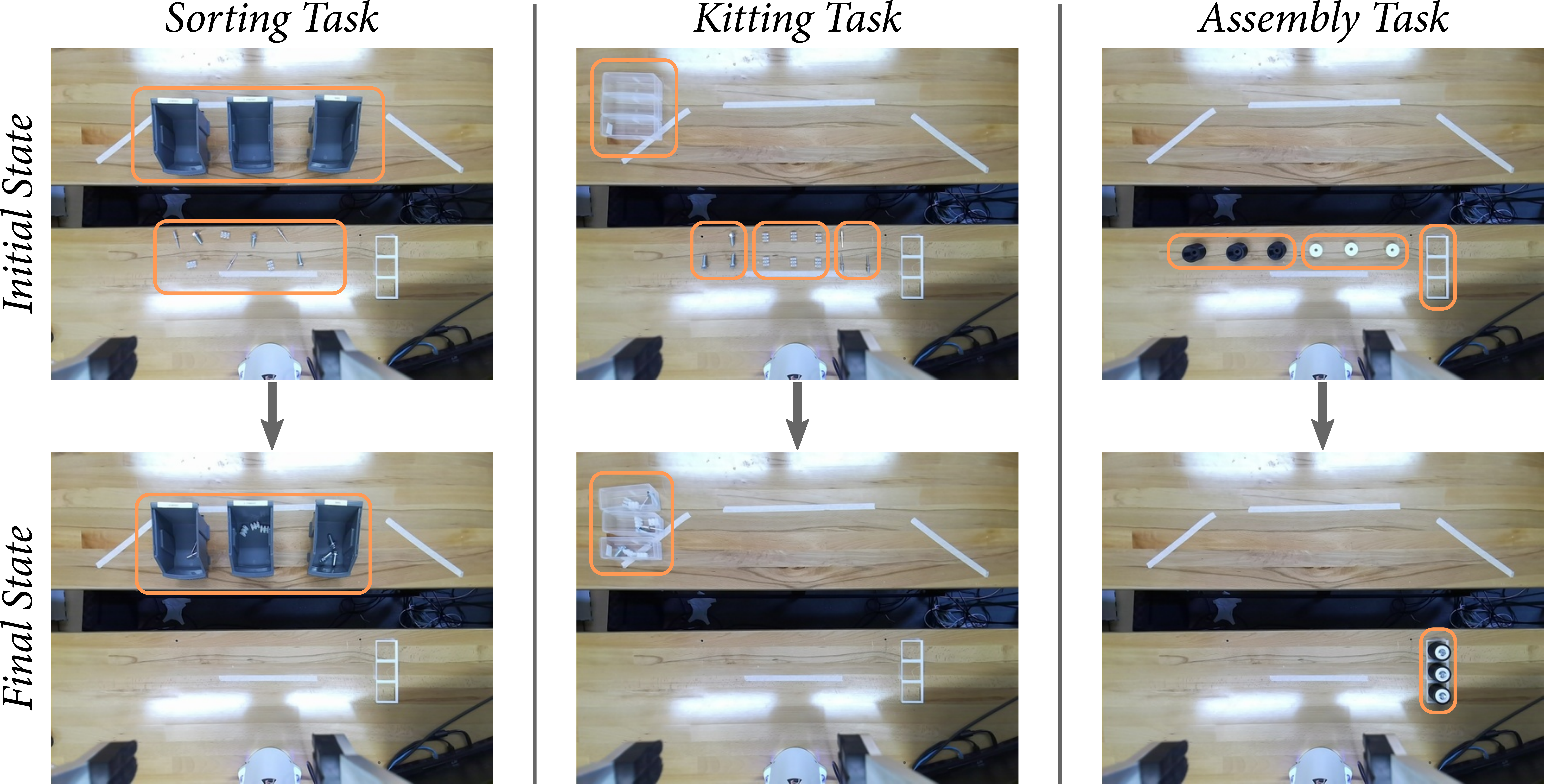}
      \Description[Picture of the initial and final states of the three tasks.]
    {
    Before and after images of each task from the robot camera in a grid with two rows and three columns.
    The initial state of the sorting task shows two highlighted rectangular zones, one around objects on the robot's table and one around three boxes on the human's table. The final state shows only one rectangle around the boxes with the objects sorted.
    The initial state of the kitting task shows four highlighted zones, one over three small transparent boxes on the human's table, and three around the three categories of objects on the robot's table. The final state only shows one highlighted zone around the three small boxes with objects inside.
    The initial state of the assembly task shows three highlighted zones on the robot's table: one around three holders, one around three top, and one around a grid structure used as a destination. The final state shows only one highlighted zone around the destination area with the places in the grid filled by the assemblies.
    }
  \caption{Example of initial and final states of the three categories of tasks, sorting, kitting, and assembly, from the robot's perspective. Orange boxes highlight task objects.}
  \label{fig:tasks}
\end{figure}

Four tasks were explored in this study: (1) sorting, (2) kitting, (3) collaborative assembly, and (4) collaborative assembly without live programming (see Figure \ref{fig:tasks}). These tasks were based on common light-manufacturing operations and were designed to have increasing levels of collaboration and to support non-unique program solutions. The order of the tasks was fixed, in order of complexity, so earlier tasks could serve as a warm up. 

During all four tasks, participants were encouraged to think aloud and voice their actions, and the experimenter could answer questions about the system (e.g., ``What does the grid pattern do''). However, the experimenter did not answer questions about how to solve the tasks (e.g., ``How can I have the robot stack object A on object B''). Additionally, the first two tasks (sorting and kitting) were used as training and served to make sure that participants understood the basic concepts required for the collaboration task where the help from the experimenter was minimal.

\paragraph{Task 1: Sorting task}
The goal of the sorting task was to sort nine objects of three different types (fasteners, bolts, and connectors) on the robot's table (bottom part of the screens on Figure \ref{fig:tasks}) into three boxes labeled with the corresponding object names on the human's table (top part of the screens on Figure \ref{fig:tasks}). The robot could detect the type and location of the objects and the boxes but did not know the labels on the boxes. The experimenter randomly changed the position of the objects and boxes between participants.

\paragraph{Task 2: Kitting task}
The goal of the kitting task was to create three kits with one bolt, two connectors, and one fastener. The objects were placed sorted on the robot's table and had to be moved to the human's table. These kits had to be put in smaller boxes that the robot could not detect. Consequently, to solve the task, participants had to design a program where the robot handed the pieces over to them so that they could take the pieces and put them in the respective boxes manually.

\paragraph{Task 3: Collaborative assembly}
The goal of the collaborative assembly task was to create three assemblies and place them on the white grid on the robot's table (bottom right part of the screens on Figure \ref{fig:tasks}), simulating a palletizing scenario. Assemblies consisted of two pieces---a \emph{top} and a bottom part (\emph{holder})---that had to be assembled and fastened with a bolt and a nut. Both the starting positions of the tops and the holders and the pallet destination were on the robot's table, while the bolts and nuts were located on the human's table. The human was the only one able to bolt the pieces together. The robot could detect the top and the holder with three pre-defined states: empty, full (when the top is stacked), and assembled (when the bolt is placed in). This task forced the human and the robot to collaborate as the robot had to act first, handing the parts over to the human and possibly stacking the top on the holder. Only then, could the human bolt the pieces together. Finally, the robot had to act again on the objects to place them on the pallet. Consequently, both agents had to react to their partner's actions.

\paragraph{Task 4: Collaborative assembly without live programming}
The goal of the collaborative assembly without live programming was to solve the same assembly task as in task 3, but this time participants were not allowed to create trigger buttons and were instructed to program all the rules at once before letting the robot move. The task order was intentionally chosen to first allow participants to use live programming in tasks 1, 2, and 3, and then disable this capability in task 4. The goal of this task was not to present a controlled comparison between live and non-live programming but to present a more challenging task to evaluate whether participants understood the logic behind trigger-action rules and were able to estimate their impact. 

\subsection{Results}
All participants were able to complete the sorting, kitting, and collaborative assembly tasks using our system. For task 3, assembly with live programming, half of the participants required a partial reset of the workspace due either to misinterpretation of the impact of an action's parameter, a missing or incorrect condition in a trigger-action pair, or incorrect expectations about the interface. Additionally, one participant asked for a reset in a situation where it was not required. For task 4, assembly without live programming, nine out of ten participants fully completed the task without having to reset the workspace. One participant required two edits of their program and one workspace reset due to distraction. One participant (P6) also solved task 3 (assembly with live programming) without using the live programming and consequently did not have to repeat their program to solve the task 4, assembly without live programming. Two other participants (P1 and P3) started with live programming, discovered an issue in their program which could require a reset and then programmed a successful behavior without using live programming. While participants were overall better at the assembly task without live programming, the study was not designed to assess whether live programming was better than non-live programming. Furthermore, there was an ordering effect as participants had already learned how the task could be solved when using the live programming.

From our analysis of participants' patterns of use and interviews, we identified three main takeaways from our study: (1) SLP supported a variety of authoring workflows; (2) SLP allowed for individualized strategies; and (3) SLP afforded a low-barrier of entry to design interactive behaviors.

\subsubsection{SLP supported a variety of authoring workflows}
The authoring process varied between the participants. In each task, participants used live programming in different ways. For example, in the sorting task, one participant (P7) created one zone for the box and a single small zone that they intentionally moved around to explicitly specify which object the robot should pick. This approach resulted in a \emph{destructive} workflow, i.e., a program that can solve the tasks, but not be reused. This use of the live programming approach was not anticipated by the authors but demonstrated how SLP allows for diverse interaction patterns. For example, it can be used in a fashion closer to teleoperation than programming.

When solving the assembly task with live programming, we observed three main ways that users adopted to tackle the problem. One participant (P6) created all three trigger-action pairs of their program at once and then simply executed it, thus not benefiting from the live programming approach. Anecdotally, this participant had significant experience in video games and described solving the task as similar to some of the problems he enjoys solving in video games. Alternatively, some participants progressively created a \emph{viable program} that would allow solving the task in future iterations if the workspace was reset. For example, participants created a first rule to move the pieces from the robot side to the human side, then while the robot moved the pieces, participants could assemble them, and finally participants programmed a second rule to move the assembled piece to the paletizing area. This strategy allowed participants to benefit from the live programming as a way to solve the task step by step and construct an autonomous program that could be reused. Finally, a last group of participants adopted a more \emph{destructive approach}, creating trigger-actions pairs when needed, and then deleting them, or moving and reusing zones once they were obsolete. For example, one participant created a first rule to stack the top parts on the holders. Once all the piece were stacked and the rule became obsolete, they deleted the rule, removed one zone, and moved the other zone to use it for their next rule. This workflow allowed the participant to solve the task while maintaining a clutter-free interface and working with a single rule at a time. However, unlike the other viable programs, if the workspace was reset, the participant would have to recreate their program from scratch.

Live programming enabled participants to see directly when some trigger-action pairs were incorrect or did not yield the expected results, allowing them to correct their program straightaway. In task 4, when not being able to use live programming, some participants appeared anxious that they were unable to modify their program during the execution. In the semi-structured interviews, most of the participants described the live programming process as being particularly useful in helping them understand how to solve the task. Participants appreciated using the robot in an exploratory fashion, as expressed in the excerpts below:
\begin{quotation}
    \textit{I definitely \textbf{liked the progressive more}, I find it very inefficient to let the whole thing run out and \textbf{then realize that there is a bug} maybe in the beginning of it. [...] I definitely prefer more \textbf{trial and error} kind of process. (P5)}
\end{quotation}
\begin{quotation}
    \textit{If I was trying to do \textbf{all the rules at once} without thinking about the task, or \textbf{without actually doing the task first}, I think that would be a \textbf{little bit overwhelming}. (P9)}
\end{quotation}

Once they had a good knowledge of how to solve the task, some participants reported that the live aspect of the programming was less important for them:
\begin{quotation}
    \textit{Since I'm more like learn what I see kind of thing, I would prefer like getting \textbf{few rules written} down, and \textbf{see how it goes and then another set of rules}, but then over time as I understand better I would just like do it all at once. (P3)}
\end{quotation}
\begin{quotation}
    \textit{I \textbf{liked doing them all at once} because I like thinking about the problem from start to finish and that was a way quicker way to do it. (P4)}
\end{quotation}
However, the same participant (P4) noted a a few minutes later in the interview: 
\begin{quotation}
    \textit{It was \textbf{helpful to do it individually first and then do them all at once}, because then I knew I was not missing any step. (P4)}
\end{quotation}

These observations matched one of the intended benefits behind the live programming approach: allowing users to get a more rapid feedback on the efficiency of their program when they are initially unsure about the output of their actions.

\subsubsection{SLP allowed for individualized strategies} 
In addition to supporting multiple authoring strategies, SLP also allowed participants to program a variety of sets of trigger-action pairs to create robot behaviors matching their mental task model and preferences. 

In task 3, \textit{assembly with live programming}, six participants created a fully autonomous robot using only conditions that the robot could trigger, while the remaining four used buttons to control the workflow. Eight out of ten participants used the robot to stack the objects first before moving them, one participant created one trigger-action pair with two actions to bring the holder and stack the top part on it, and one participant moved all the holders prior to stacking the top parts. 

For the \textit{collaborative assembly without live programming}, seven participants had the robot stack the pieces before moving the partially assembled piece. However, even if the rules were similar, the resulting workflow could be different due to the priority selection and variations in the exact conditions used. For example, during the interview some participants reported that they tried to optimize the interaction, for example by deciding which tasks to assign to the robot, organizing zones in the workspace, or by prioritizing trigger-action pairs to increase the parallelization opportunities:
\begin{quotation}
    \textit{I was trying to think kind of the \textbf{most efficient}, where you have, say, me as an operator doing one task, the \textbf{robot should be doing another task}. (P1)}
\end{quotation}
\begin{quotation}
    \textit{I did notice like when I needed to assemble it myself it was nice \textbf{having the robot doing the other thing}, so I could see how it would be helpful. (P4)}
\end{quotation}
\begin{quotation}
    \textit{I tried to see \textbf{what the robot could do} and \textbf{what I could do} and which one was the \textbf{easiest option}, and I kind of figure out the most, the \textbf{fastest way}, the \textbf{most efficient way}. (P7)} 
\end{quotation}

These observations show that SLP allowed participants to design interaction flows suited to their personal preferences, which demonstrates the flexibility offered by this approach.

\subsubsection{SLP afforded a low-barrier of entry to design interactive behaviors}
Participants evaluated positively the usability of our system (SUS: $M(SD)$=$75.8(19)$) which according to \cite{brooke2013sus} indicates a usability between ``good'' and ``excellent''. Notably, one of our participants (P8) had more issues with the system than the others. The participant was confused with the impact of some conditions, for example ``is not in'' could both be a boolean checking the absence of an object in the zone or a filter referring to all the objects not in the zone. When facing such a ambiguous situation, participants first inspected their rules, trying to identify what was the origin of the unexpected behavior, and then had to update their trigger-action pair to find another valid way to solve the task.

These high usability scores were supported by participants answers in the semi-structured interview:
\begin{quotation}
    \textit{The button interface was intelligent and \textbf{easy to use}. (P2)}
\end{quotation}
\begin{quotation}
    \textit{I thought it was \textbf{easy to use}, \textbf{easy to learn}. (P6)}
\end{quotation}
Participants reported that it was easy to use the \textit{zones} to refer to objects and noted the importance of having access to the state of the objects in the conditions:
\begin{quotation}
    \textit{I think this is a lot easier [than other programming methods] because there is really \textbf{good visuals} and \textbf{you're seeing exactly what you're doing}, that's easier to think about it because you can really visualize what you're doing. (P9)}
\end{quotation}
\begin{quotation}
    \textit{The fact that it could \textbf{recognize whether a part was just filled or assembled} made it a lot easier to use. (P2)}
\end{quotation}
Participants also reported that the trigger-action programming framing for the collaboration was intuitive, easy, and natural:
\begin{quotation}
    \textit{I liked the setup how you had the statement like if it is in the zone, then you know with \textbf{the condition and the action}, I thought that it \textbf{made a lot of sense}. (P9)}
\end{quotation}

A few noted that even though no expertise in programming was required, computational thinking was useful:
\begin{quotation}
    \textit{It wasn't directly programming, so \textbf{I don't think you need a lot of programming experience}, it's just more, knowing \textbf{how computer logic works} is definitely very useful for creating that. (P1)}
\end{quotation}
\begin{quotation}
    \textit{I coded a bit back a year ago or something, over time with the human-robot interaction could like brush back those knowledge and apply it, \textbf{overtime it became seamless}. (P3)}
\end{quotation}
Participants also reported that it often took them some time to understand the robot's capabilities, but once they learned how to program the robot, they were confident in knowing what the robot could do and how to do it:
\begin{quotation}
    \textit{I think once I did the first task it was a lot easier to know the capabilities of the robot moving forward. And then, therefore, \textbf{knowing these capabilities it was easier for me to make the more complex rules} with the multiple conditions. (P5)}
\end{quotation}
\begin{quotation}
    \textit{It was \textbf{initially a little confusing} a little, like figuring how what each thing does and what control the robot, but \textbf{by the end I felt pretty confident}. (P7)}
\end{quotation} 

Participants proposed a number of points to improve the interface and the system. For example, participants desired more direct feedback from the interface such as visual feedback on the impact of action (P1), verification for errors (P5), identification of actions with a risk to fail (P9), and anticipation of priority conflicts (P6).

\section{Discussion}
In this section, we describe takeaways from our study results, limitations of SLP, our implementation, and our evaluation as well as a roadmap to extend SLP to improve its applicability to a broader range of human-robot collaboration scenarios.

\subsection{Observations}

Our exploratory user study showed that participants with some programming experience could easily learn and use our interface to create interactive behaviors for human-robot collaboration. Furthermore, SLP allowed participants to leverage live programming to solve collaborative problems in various ways. Examples include deleting obsolete rules and zones to maintain a less cluttered interface and a simpler program or waiting to see the outcome of a trigger-action pair before starting to program the next one. These additional workflows would not be supported by traditional programming where the robot program is expected to solve the full task directly.
Participants reported that our system was intuitive, but they also desired to have more feedback from the interface and the robot. For example, one participant reported that it would be useful if the robot could communicate when it is about to execute an action via visual or audio feedback. Some participants noted that it would be beneficial for the interface to verify the correctness of their trigger-action pairs or help them to debug programs. Furthermore, as SLP combines multiple trigger-action pairs that can interact with each other, future interfaces could add symbolic checking, for example, by evaluating if one trigger-action pair could trigger another one or predicting priority conflicts. Formal verification has a long history in robotics \cite{luckcuck2019formal}, and recent work in end-user authoring tools explored how verification methods could be applied to end-user authored programs \cite{porfirio2018authoring}. A similar approach could benefit SLP. Participants also requested rapid feedback regarding the impact of the parameters directly on the interface, for example, in the form of displaying the destination of objects corresponding to a move action. This addition would result in a second layer of liveliness that is more akin to classical live programming \cite{tanimoto2013perspective}.

To make trigger-action pairs more concise and easier to program, our system used the conditions within the trigger to set a context for the actions. However, this context usage resulted in confusion for some participants. One participant was confused when creating a trigger-action: ``if holder is not in yellow zone then move holder from green zone to yellow zone.'' The participant interpreted this condition as being ``for each holder not in yellow zone'' whereas the system interpreted it as ``if there is no holder in yellow zone.'' Our syntax made the trigger ambiguous. A more general solution might involve requiring the rules to be fully explicit (i.e., not simplifying based on context) or more carefully crafting the language of the rules to avoid confusion.

\subsection{Limitations \& Future Extensions}

We believe that SLP can benefit a wide range of tasks and applications within human-robot collaboration. In this work, we focused on introducing the concept and demonstrating basic examples of how this paradigm enables programming of collaborative task actions for robots. However, our SLP approach, the current implementation, and the evaluation contain a number of limitations. In this section, we discuss these limitations and describe a roadmap for how these limitations can be addressed in future extensions.

\subsubsection{Limitations of the exploratory study}
While demonstrating the opportunity to use SLP to program robots for collaboration, our study had a number of limitations that limit the applicability of our study beyond the explored environment. First, participants were drawn from a student population and not factory workers or domain experts such as carers. To assess the applicability of SLP to specific environments, future work should explore whether the actual end users, e.g., factory workers, could use SLP to solve tasks they face in practice. Second, as the study was exploratory, it did not compare SLP to other methods (e.g., programming by demonstration, teach pendants, or other authoring methods), neither did it assess the importance of the situated and the live aspect of the method. Future work should both compare SLP to other methods, and assess through rigorous ablation studies the impact of each aspect of SLP. Third, the study only explored simple tasks, with only one robot and one human. Future work should evaluate SLP in more complex environments, beyond tabletop manipulation, with more realistic collaboration problems, more actions, and with more diverse human-robot teams. 

\subsubsection{Limitations of the SLP Approach}
While holding promise for programming robots for collaboration, SLP also has some limitations. First, unlike many scheduling or planning approaches, SLP does not offer guarantees in completeness or optimality. By relying on the human collaborator to specify the robot program, SLP is bounded by human capabilities. To promote a more exploratory approach to programming, SLP does not prevent users from making errors (e.g., stacking two objects by accident). However, in non-critical environments, the user can reset the workspace to a previous state and correct their program. Second, similar to other end-users programming tools \cite[e.g.,][]{gao2019pati}, our approach only allows the use of high-level actions for specification, which limits the program to tasks that can be specified using only those predefined high-level actions. To extend the capabilities of these tools, future extensions can provide users with the ability to create high-level actions, use programming by demonstration to parameterize actions, or specify templates that can be manipulated with the interface. 

\subsubsection{Generalization}
Application to new domains or tasks could require detection of new objects and specification of additional action primitives (e.g., ``insert'' or ``press''). Similar to other task-level authoring tools \cite[e.g.,][]{paxton2017costar,gao2019pati}, SLP relies on reliable object detection and proper action handling. We envision three ways to enable the system to adapt to situations involving new objects. First, a batch-processing mode can provide the system with a set of expected objects for a given context. For example, before introducing it to the floor of a machine shop, the system can be trained using a set of models of typical hardware (e.g., bolts, fasteners) and tools (e.g., molds, power tools) used in the shop or using synthetic datasets \cite{mccormac2017scenenet} for a large number of hardware and tools used in the industry. Second, new objects could be added and trained on the fly. In this case, a user could identify an unrecognized object and add it to the library by reorienting the object and capturing several identifying images using the the robot's end-effector camera \cite{wang2019hand}. Finally, for objects with infrequent or irregular use, primitive ways to identify objects, such as specifying grasp locations directly through the interface \cite{kent2020leveraging} could facilitate action without the overhead of training an object model. However, when adding new objects to the system, the user would also have to specify the available affordances (possible states, actions) to be able to use them in the interface. Additionally, we only used visual ways to evaluate the state of objects, and future work could explore physical checks, such as the torque used to tighten a bolt, as a way to evaluate the state of the environment.

Certain operations, such as retasking the robot to a different process, could also require novel actions. The current system assembles actions using a series of primitives (for example, the ``move to'' action consists of eight ``move'' and two ``toggle gripper'' primitives) which allows modular creation of new actions. The system presented in this paper supported developers in adding new actions, objects, and states, but not end users. To be deployed as an end-user programming tool, the tool must provide end users with the ability to define new actions. A large body of work in the human-robot collaboration literature focuses on the creation of new skills, for example, using learning from demonstration \cite{billard2008survey}, by combining primitives using an interface such as RAFCON \cite{brunner2016rafcon} or COSTAR \cite{paxton2017costar}, or by combining multiple modalities to specify actions such as CODE3 \cite{huang2017code3}. A future system could combine such methods to create new high-level actions and integrate them in our SLP authoring system and interface to generalize beyond the environment studied in this paper. 

Finally, similarly to any high-level authoring system \cite[e.g.,][]{gao2019pati,steinmetz2018razer}, our system relies on robust actions with a low probability of failure. However, to be deployed outside of the lab, systems will need to handle errors more robustly. With SLP, error handling could be approached at three levels: the primitive level (e.g., a failed grasp could be detected and tried again), the action level (e.g., using motion planning a fail grasp could lead to a replanning of the grasp pose), and at the TAP level (e.g., additional rules could be created to handle the failure of other rules). Future work should explore the usability and robustness of these different error handling levels.

Allowing users to introduce new objects to the system, new ways to evaluate the state of the environment, create new actions on the fly, and handle manipulation errors would greatly increase the real-world applicability of our approach. With these improvements, workers would be able both to increase the skill repertoire of the robot and improve the way such skills could be used.

\subsubsection{Environment Complexity}
In our current implementation, the workspace is contained within a single two-dimensional view, and tasks are void of excessive clutter. Certain environments may require actions spanning multiple views or multiple non-collocated workstations. For example, a mobile cobot in a hospital may need to travel between an exam room and a supply closet to retrieve a set of supplies needed for a given procedure. Alternatively, in manufacturing, a robot might have to pick objects from a conveyor belt and move them to a pallet. Such scenarios require interactions with a larger workspace that may not fit in a single view. Our system allowed users to control of the camera position and orientation to view other parts of the workspace and offered additional actions, including push, pull, tighten, and loosen, but these functions were disabled as they were not critical for the evaluation. Future work should explore how the interface could allow the user to navigate within the environment (both around the robot and in different locations) and address this more complex visualization problem. For example, systems could allow users to \emph{bookmark} important view points for the robot at the start or during the live programming process and create triggers that the robot responds to by moving between these viewpoints, enabling the robot to react to state changes in more than one workspace without relying on external sensors. 

When environments become cluttered or complex, execution of actions can be complicated. For example, unpacking a box requires complex motion planning to accomplish the action without collisions. To enable more robust execution in complex environments, we plan to integrate state-of-the-art motion planning and interactive task planning to allow the end user to provide interventions or corrections when a simple action is insufficient to complete the task \cite{hagenow2021corrective}. For example, if the robot fails to detect obstacles during execution of an action, the collaborator could intervene by kinesthetically guiding the robot around the obstacle or by manipulating a visual representation of the path, for example in a mixed-reality environment \cite{ostanin2020human}.

Finally, more complex tasks may require programs composed of a large set of complex rules. The increase of rule complexity and number of rules will raise questions about transparency and debuggability. To address these issues, a clearer syntax needs to be developed. Future work should investigate how to better communicate what objects would be used in actions, how to express the rules in a succinct, yet transparent way, and how to maintain understandability and debuggability as programs grow more complex. Furthermore, future work should explore how complex priority cases and interruptions should be handled by the system, for example by specifying the criticality and interruptibility of each trigger-action pair.

\subsubsection{Multi-agent Collaboration}
For certain tasks, it may be beneficial to employ multiple robotic agents to assist one human collaborator. For example, it may be desirable to have a general purpose manipulator for assembly tasks and a second mobile agent responsible for moving objects around a warehouse. Although SLP supports interaction between multiple robotic agents using trigger-action pairs, additional safety mechanisms would be necessary to avoid collisions in the workspace and with the human collaborator. These mechanisms could include real-time motion planning and collision avoidance as well as the use of active compliance on robotic manipulators to avoid injury or damage during unintentional collisions.

\section{Conclusion}

With this paper, we propose to use situated live programming for HRC by framing the robot programming within the TAP paradigm and supporting the incremental programming of rules for collaborative tasks. Using multiple trigger-action pairs to define a program inherently handles interaction, allows for easy extension of the program at runtime, and provides new workflows for users to solve tasks. This live programming can further be situated in the environment by using an augmented video feed as background for authoring. With our study, we demonstrated that SLP is a viable framework for HRC: it allows users with little programming experience to create interactive behaviors to solve simple collaboration task and many of these users made use of the incremental programming opportunity offered by SLP. We believe that this approach to robot programming for collaboration, by shifting the focus of robot programming from a pre-deployment process to a situated and incremental one, extends the toolbox available to designers and developers to allow end users to retask robots in situations where behaviors are highly contextualized in the current state of the world, should be created by novice users, and might be benefit from being extended or edited at runtime.

\begin{acks}
This work was funded by a NASA University Leadership Initiative (ULI) grant awarded to the University of Wisconsin--Madison and The Boeing Company (Cooperative Agreement \#80NSSC19M0124). We would also like to extend our gratitude to Titus Smith who developed the interface.
\end{acks}

\bibliographystyle{ACM-Reference-Format}
\bibliography{paper}


\end{document}